\documentclass[reprint, prl, aps, superscriptaddress, floatfix]{revtex4-2}

\usepackage{hyperref}
\hypersetup{
    colorlinks=true,
    linkcolor=blue,
    citecolor=red,
    urlcolor=blue,
    pdftitle={MaskTerial},
    pdfpagemode=FullScreen,
    }
\usepackage{graphicx}
\usepackage{amsmath}
\usepackage{amsfonts}
\usepackage{color}
\usepackage{verbatim}
\usepackage{upgreek}
\usepackage{bm}
\usepackage{textcomp}
\usepackage[symbol*]{footmisc}
\usepackage{float}
\usepackage{xcolor}
\usepackage{multirow}
\usepackage{tabularx}
\usepackage{booktabs}
\usepackage{amssymb}
\usepackage{pifont}
\newcommand{\cmark}{\ding{51}}
\newcommand{\xmark}{\ding{55}}

\DefineFNsymbolsTM{otherfnsymbols}{
  \textdagger    \dagger
  \textdaggerdbl \ddagger
  \textsection   \mathsection
  \textparagraph \mathparagraph
 }
\setfnsymbol{otherfnsymbols}

\newcommand{\beq}{\begin{equation}}
\newcommand{\eeq}{\end{equation}}
\newcommand{\bea}{\begin{eqnarray}}
\newcommand{\eea}{\end{eqnarray}}
\newcommand{\bal}{\begin{align}}
\newcommand{\eal}{\end{align}}

\begin{document}
\title{MaskTerial: A Foundation Model for Automated 2D Material Flake Detection}

\author{Jan-Lucas Uslu}
\email{jan-lucas.uslu@rwth-aachen.de}
\affiliation{2nd Institute of Physics and JARA-FIT, RWTH Aachen University, 52074 Aachen, Germany}
\affiliation{Visual Computing Institute, RWTH Aachen University, 52074 Aachen, Germany}
\author{Alexey Nekrasov}
\author{Alexander Hermans}
\affiliation{Visual Computing Institute, RWTH Aachen University, 52074 Aachen, Germany}
\author{Bernd Beschoten}
\affiliation{2nd Institute of Physics and JARA-FIT, RWTH Aachen University, 52074 Aachen, Germany}
\author{Bastian Leibe}
\affiliation{Visual Computing Institute, RWTH Aachen University, 52074 Aachen, Germany}
\author{Lutz Waldecker}
\email{waldecker@physik.rwth-aachen.de}
\affiliation{2nd Institute of Physics and JARA-FIT, RWTH Aachen University, 52074 Aachen, Germany}
\author{Christoph Stampfer}
\affiliation{2nd Institute of Physics and JARA-FIT, RWTH Aachen University, 52074 Aachen, Germany}
\affiliation{Peter Gr\"unberg Institute (PGI-9) Forschungszentrum J\"ulich, 52425 J\"ulich, Germany}

\begin{abstract}
The detection and classification of exfoliated two-dimensional (2D) material flakes from optical microscope images can be automated using computer vision algorithms.
This has the potential to increase the accuracy and objectivity of classification and the efficiency of sample fabrication, and it allows for large-scale data collection.
Existing algorithms often exhibit challenges in identifying low-contrast materials and typically require large amounts of training data.
Here, we present a deep learning model, called MaskTerial, that uses an instance segmentation network to reliably identify 2D material flakes.
The model is extensively pre-trained using a synthetic data generator, that generates realistic microscopy images from unlabeled data.
This results in a model that can to quickly adapt to new materials with as little as 5 to 10 images.
Furthermore, an uncertainty estimation model is used to finally classify the predictions based on optical contrast.
We evaluate our method on eight different datasets comprising five different 2D materials and demonstrate significant improvements over existing techniques in the detection of low-contrast materials such as hexagonal boron nitride.
\end{abstract}

\maketitle
\date{\today}

\section{Introduction}

The ability to combine different 2D materials into van der Waals heterostructures has opened up new ways to study fundamental phenomena in solids~\cite{Novoselov2005Jul, Geim2007Mar, Novoselov2016Jul, Zhong2017,Cao2018,Niu2022}, to tailor material properties~\cite{Britnell2013May,Ferrari2015, Androulidakis2018, Fang2019, Tebbe2023, Volmer2023Aug} and to design device structures with improved performance~\cite{Koppens2014, Mennel2020, Sierra2021, Lemme2022, Icking2024}.
In most research settings, these heterostructures are assembled from individually exfoliated material flakes~\cite{Yi2015,Frisenda2018}.
The identification and selection of suitable 2D material flakes for device fabrication is the first and an integral part of this process~\cite{Uslu2024}.
It has traditionally been performed by researchers who scanned large pieces of exfoliation substrates using a microscope.

\begin{figure}[!ht]
    \begin{center}
        \includegraphics
        [width = \linewidth]{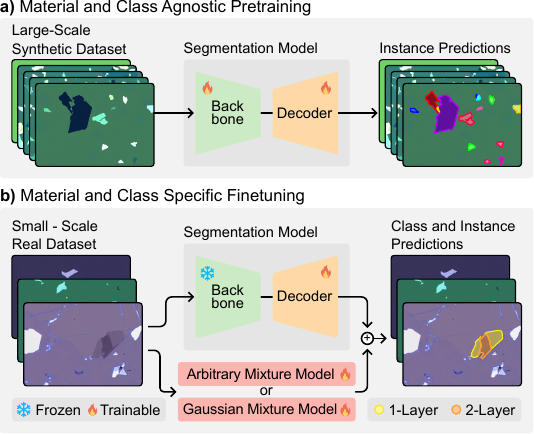}
        \caption{
MaskTerial uses a two step approach to train a robust foundation model for 2D material flake detection.
\textbf{(a)}
The segmentation model is pre-trained on a large corpus of synthetic data of multiple to learn a good internal representation of the data.
The synthetic data contains no information about the material type or thickness of the flakes.
This results in a material and class agnostic pre-trained foundation model.
During this step, both the backbone and the decoder are trained.
\textbf{(b)}
After pre-training, a small number of images is used to fine-tune the classification model and the decoder of the foundation model.
The classification model can be either the new arbitrary mixture model or any existing Gaussian mixture model from ref.~\cite{Uslu2024}.
}
        \label{fig:Hook_Figure}
    \end{center}
\end{figure}

Automating the detection of exfoliated 2D material flakes using computer vision algorithms has the potential to significantly improve sample preparation efficiency and accelerate the pace of research~\cite{Ryu2022}.
For this task, previous work has explored the use of classical machine learning methods, such as support vector machines (SVMs) and K-means clustering~\cite{Masubuchi2018, Lin2018, Li2019}.
These methods rely on the discrete nature of the optical contrast values of 2D materials with respect to the substrate material~\cite{Blake2007}.
This discrete nature is a result of their atomic-scale thickness, where each layer of material corresponds to a single, uniform atomic plane.
Optical contrast variations arise due to interference effects, where the interaction of light with the material and substrate depends on the exact number of these atomic layers, leading to quantized optical contrasts for each layer count.
However, the performance of current detection models typically decreases significantly for materials with low optical contrast and they are sensitive to variations in substrate thickness and lighting conditions~\cite{Uslu2024}.
More recently, deep learning approaches using neural networks have been employed to address these limitations~\cite{Saito2019, Han2020, Masubuchi2020}.
Although these methods offer greater versatility, they typically require large amounts of labeled training data, which can be impractical to obtain in a research setting, especially if the yield of exfoliated materials is low.

Recently, the emergence of foundation models in artificial intelligence has transformed numerous fields by providing pre-trained models that can be fine-tuned for diverse tasks with minimal labeled data~\cite{Bommasani2021}.
Foundation models, such as GPTs~\cite{Brown2020} for language and vision transformers~\cite{Dosovitskiy2021} (ViTs) for image processing, leverage extensive pre-training on large and diverse datasets, enabling them to generalize across domains with limited additional training.
These models are often trained on large-scale datasets using self-supervised or unsupervised learning techniques, allowing them to capture broad representations of data.
This versatility makes them particularly powerful for tasks where labeled data is scarce or hard to obtain, as they can transfer learned features effectively to new domains.

The success of foundation models in other domains inspires the potential for similar advancements in 2D material flake detection.
By leveraging pre-trained models and domain-specific fine-tuning, foundation models can address key limitations such as the need for large labeled datasets and the challenges posed by low-contrast materials.
Building on these principles, we propose a tailored approach to tackle the specific challenges of 2D material flake detection.
First, we introduce a deep learning architecture that combines a modified Mask2Former~\cite{Cheng2022} model for instance segmentation with a physics-informed uncertainty estimation head based on the deep deterministic uncertainty (DDU) method~\cite{Mukhoti2023}.
This architecture allows for robust detection and classification of 2D material flakes, even for low-contrast materials such as thin hexagonal boron nitride (hBN).
Second, we propose a synthetic data generation pipeline using physical simulations in conjunction with unlabeled data to address the lack of large amounts of labeled data.
We show that extensive pre-training of our model using synthetic data (see Figure \ref{fig:Hook_Figure}a) allows it to be fine-tuned with as few as 5 to 10 microscope images per material (see Figure \ref{fig:Hook_Figure}b).
Finally, we present eight different datasets covering five different 2D materials used for training and evaluation to validate the performance of our models.

\begin{figure}[!th]
    \begin{center}
        \includegraphics[width=\linewidth]{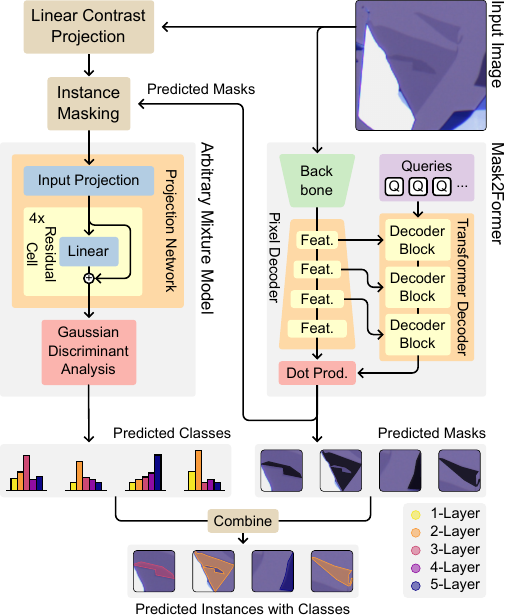}
        \caption{
MaskTerial consists of two models: an instance segmentation model and a classification model.
First, the input image is processed by the instance segmentation model returning a set of possible flakes without classifying their thickness class.
If possible flakes are found, the image is projected into the contrast space representation following the method described in ref.~\cite{Uslu2024}.
Afterwards, the masks of the predicted instances are used to extract the contrast values from the transformed image for each of the instances.
These are then classified by the classification model to generate probability distributions over the classes of the flakes.
The mode of these distributions is used to classify each instance, yielding the final predicted flakes with thickness classes.
}
        \label{fig:model_architecture}
    \end{center}
\end{figure}

\section{Model Architecture}

The MaskTerial architecture combines two deep learning models.
The first model, the instance prediction model (Figure \ref{fig:model_architecture} - Mask2Former), predicts all flakes of interest in the image, regardless of the actual class of the predicted flake.
The second model (Figure \ref{fig:model_architecture} - arbitrary mixture model) then takes all the predicted interesting flakes and assigns them classes based on their contrasts, i.e. monolayer, bilayer, etc.
This separation of instance prediction and class prediction has the benefit that, when adding new materials, only the latter model needs to be retrained.

\begin{figure*}[!bth]
    \begin{center}
    \includegraphics[width=\linewidth]{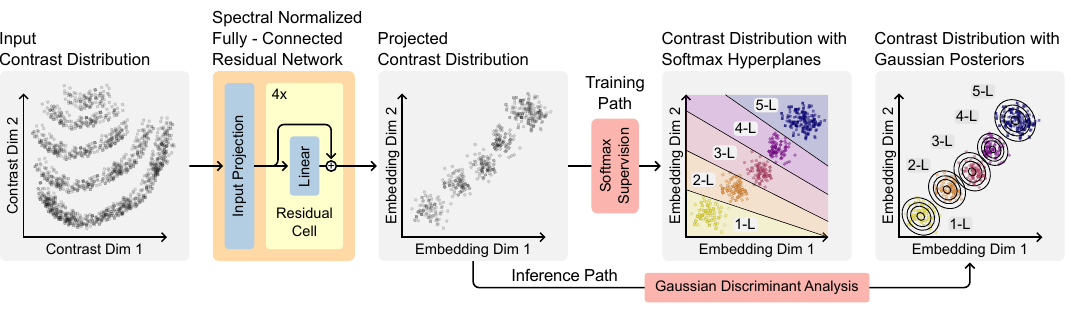}
        \caption{
The arbitrary mixture model (AMM) works by projecting an input distributions onto a distribution of Gaussians.
This is achieved by combining a ResNet with spectral normalization techniques.
The combination of regularization and residual connections allows the training process to be supervised using a straightforward softmax function and standard cross-entropy loss.
After training, a Gaussian is fitted to the embedding representations of each class.
During inference, the embedding representations of the input data are evaluated against the learned Gaussians from training to determine class conditional posteriors and thus the probability that any given input contrast belongs to a given thickness class.
        }
        \label{fig:AMM}
    \end{center}
\end{figure*}

\subsection{Instance Prediction Model}

The instance prediction model is based on the Mask2Former~\cite{Cheng2022} architecture (see Figure \ref{fig:model_architecture}).
It works by first extracting feature representations from the input image using a ResNet50~\cite{He2016} backbone.
Afterwards, the extracted feature representations are gradually upscaled by a pixel decoder (PD).
During upscaling, the features are sequentially fed into the transformer decoder (TD) at multiple levels.

A unique aspect of Mask2Former is its use of learnable query embeddings, introduced by the DETR~\cite{Carion2020} architecture.
We train these queries to act as proxies for potential object instances or specific semantic categories within the image.
During the decoding process, these queries interact with the encoded image features via cross-attention mechanisms within the TD contextualizing the queries.
These contextualized queries are then used to generate segmentation masks of each object by computing the dot product between them and the final feature representation from the PD.
In our case, the segmentation does not classify the instance by layer count, such as monolayer or multilayer; instead, it only identifies interesting objects (i.e. a 2D material flake).
This improves the detection accuracy for downstream tasks (see Table \ref{tab:ablations}).

\subsection{Classification model}

The second component of MaskTerial is the arbitrary mixture model (AMM), which assigns layer thicknesses to optical contrasts of the flakes.
As discussed in our previous work~\cite{Uslu2024}, variations in the oxide thickness in the Si/SiO$_2$ wafers, used to exfoliate the 2D material flakes, lead to non-trivial distributions of these contrasts (see Figure \ref{fig:AMM} - Input Contrast Distribution), making them difficult to fit and the detection unreliable.
To counteract this, we propose a model which learns a regularized mapping of arbitrary class distributions in the optical contrast space to Gaussians, solving the problem of non-trivial distributions while preserving the interpretability of the contrast distributions (see Figure \ref{fig:AMM} - Projected Contrast Distribution)).
To achieve this, we use an approach for uncertainty estimation in deep learning proposed by Mukhoti et al.~\cite{Mukhoti2023}, who introduced the deep deterministic uncertainty (DDU) method.
It addresses the limitations of traditional probabilistic models, such as Bayesian neural networks, which can be computationally expensive and difficult to train~\cite{Chandra2019}.

They proposed to use a combination of spectral normalization and residual connections to constrain the model to learn a smooth and locally linear embedding space.
Spectral normalization works by constraining the eigenvalues of each weight matrix by dividing them by their largest eigenvalue during each training step~\cite{Miyato2018}.
Residual connections allow the network to learn perturbations around the identity function, which has been shown to improve the stability and convergence of deep networks~\cite{He2016}.
Together, these two methods impose a bi-Lipschitz constraint on the model, leading to a robust and sensitive embedding space while preventing feature collapse to a single point~\cite{Mukhoti2023, Miyato2018, Liu2020}.

During inference, the class probabilities are computed by evaluating the probability density function of each class-conditional Gaussian at the embedding space coordinates of the input feature, giving a probability of the instance belonging to any given class.

The ability to interpret the resulting distributions is an important aspect of training the model in this way.
Unlike typical deep learning models that learn arbitrary functions minimizing some objective function, the model can provide uncertainty estimates based on a regularized projection to class-conditional Gaussians.

\section{Synthetic Data Generation}

Effective training of deep learning models requires a large amount of labeled data to ensure that the model accurately captures the underlying data distribution~\cite{DeepLearning}.
However, collecting and annotating real-world data is often challenging and time consuming.
To address this issue, we developed a synthetic data generation engine that incorporates physical knowledge and simulations to generate images that closely resemble real-world microscopy images, a technique often used to train self-driving cars~\cite{Richter2016, Ros2016}.

\begin{figure}[!t]
    \begin{center}
    \includegraphics[width=\linewidth]{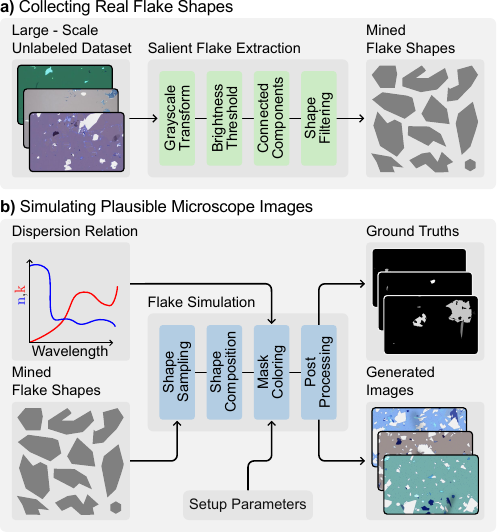}
        \caption{
The workflow for generating synthetic flake images consists of two steps.
\textbf{a)} The process begins by mining real flake shapes from a large unlabeled dataset.
This involves first converting images to grayscale, applying brightness thresholds, finding connected components in the resulting binary masks, and finally filtering the detected shapes to keep only high quality flake shapes.
\textbf{b)} The second step uses the previously mined flake shapes to generate plausible synthetic images with associated ground truth masks.
This is done by first sampling a set of shapes from the mined shapes and scattering them over an empty placeholder image to create a grayscale ground truth image.
Then, using the optical dispersion relation of the target material and the setup parameters such as visible light spectrum, camera activation curve, and substrate thickness, the colors of the material are simulated.
Finally, post-processing, such as adding noise, vignetting and shadows, is applied to create the final synthetic image.
}
        \label{fig:synth_generation}
    \end{center}
\end{figure}

The image generation process has two main phases, a shape mining phase in which we extract plausible flake shapes from a dataset of unlabeled images (Figure \ref{fig:synth_generation} a), and a generation phase, in which we use the extracted shapes to generate new images together with the ground truth masks (Figure \ref{fig:synth_generation} b).

We extract shapes from a dataset of around 100,000 unlabeled images of exfoliated graphite from an internal database.
Since most commonly used 2D materials have a hexagonal crystal structure, we assume that the shapes of their exfoliated flakes will generally be similar to those of graphite.

The images are converted to grayscale, and we then apply a stepped brightness threshold.
By setting specific brightness ranges to one and all other values to zero, we create binary masks for different brightness levels.
We then use a connected components algorithm~\cite{Bolelli2020} to extract all connected shapes from the binary masks.
Finally, we filter these shapes using an L2 classifier~\cite{Uslu2024}.
In total, 35,000 flake shapes were collected.

In the image generation process (Figure \ref{fig:synth_generation} b), we sample a random number of shapes (ranging from 1 to 500) and randomly place them on an empty canvas at different angles, sizes, positions, and thicknesses, creating a grayscale image where the pixel values correspond to the number of layers for any given pixel.
When shapes overlap, their layer counts are added in the overlapping area, creating a grayscale mask.

The color of the flakes and the background is approximated using a simulation based on the transfer-matrix method (TMM).
First, the reflectance of each pixel is calculated taking into account the thickness of the SiO$_2$ layer of the substrate, the dispersion relation of the material and the thickness of the pixel considering the grayscale mask.
The color is calculated by integrating the simulated spectral reflectance multiplied by the camera activation curve and the light source spectrum for each RGB channel.

Finally, the images are post-processed by adding a layer of random tape residue emulated with simplex noise, random shadows, a vignetting effect, and Gaussian camera noise to closely resemble real images.

We generated about 42,000 synthetic images with ground truth masks each for graphene, chromium triiodide (CrI$_3$), hBN, tantalum disulfide (TaS$_2$), molybdenum diselenide (MoSe$_2$), tungsten disulfide (WS$_2$) and tungsten diselenide (WSe$_2$), resulting in a total of about 300,000 synthetic images, which where used for pre-training.

\section{New Datasets}

\begin{table}[]
    \renewcommand{\arraystretch}{1.4}
    \centering
    \begin{tabularx}{\linewidth}{
    >{\centering\arraybackslash}X
   >{\centering\arraybackslash}X
   >{\centering\arraybackslash}X
  >{\centering\arraybackslash}X }
    \toprule
    \multirow{2}{*}{Dataset} & Train/Test Images & Annotated Classes & Train/Test Exfoliations \\
    \midrule
    Graphene (Low)      & \multirow{2}{*}{425/1362}&   \multirow{2}{*}{1-4 Layers} & \multirow{2}{*}{2/10}\\
    Graphene (Medium)      & \multirow{2}{*}{357/325} &   \multirow{2}{*}{1-4 Layers}& \multirow{2}{*}{8/9}\\
    Graphene (High)      & \multirow{2}{*}{438/480} &   \multirow{2}{*}{1-4 Layers}& \multirow{2}{*}{10/10}\\
    \midrule
    WSe$_2$ (Low)       & 92/420  & 1-3 Layers & 2/12 \\
    WSe$_2$         & 97/99   & 1-3 Layers & 5/5\\
    \midrule
    hBN             & 73/62   & 1-3 Layers & 2/3\\
    WS$_2$          & 53/94   & 1 Layer & 2/2\\
    MoSe$_2$        & 63/97   & 1-2 Layers & 7/8\\

    \bottomrule
    \end{tabularx}
\caption{
We collected eight datasets from five materials to measure the performance of the models on different materials and substrate variations.
We chose a 50/50 train test split to better capture the data distribution in the test set.
}

\label{tab:datasets}
\end{table}

To fine-tune and evaluate the model, we collected eight new datasets from five different materials.
We collected three datasets for exfoliated graphite and two datasets for WSe$_2$ with different substrate thicknesses to measure the robustness of the models.
These datasets are the low, medium, and high variance datasets to denote the range of different substrate thicknesses in the training and testing sets.
The low variance datasets contain images with substrate thicknesses within $\sim5$nm of the $\sim90$nm substrate thicknesses used.
The medium and high variance datasets contain images with ranges of $\sim10$nm and $\sim20$nm, respectively.
In addition, we have collected datasets for hBN, MoSe$_2$, and WS$_2$ with substrate thickness variations of about 10 nm.
The training and test images are from independent exfoliation runs to ensure that the test images do not bleed into the training images.
Table \ref{tab:datasets} lists the datasets, the number of images in the train and test sets, and the number of exfoliation runs in the train and test sets.

\section{Training}

\subsection{Instance Prediction Model}

The instance prediction model was trained in two stages.
First, we performed extensive pre-training using the 300,000 simulated images on 8 NVIDIA V100 GPUs, see Figure \ref{fig:Hook_Figure}a).
We trained for 90,000 iterations with a batch size of 56 with images cropped to a resolution of  $1024 \times 1024$.
We used the AdamW~\cite{loshchilov2018} optimizer with a learning rate of $10^{-4}$, a weight decay of $5 \cdot 10^{-2}$.
The learning rate scheduler we used was a simple linear decay scheduler.
Finally, we enabled gradient clipping throughout the model and clipped them to $10^{-2}$.
The pre-training took about 52 hours.

For further fine-tuning, we used the pre-trained instance prediction model as a base, using the same parameters and number of images for all materials while freezing the parameters of the backbone, see Figure \ref{fig:Hook_Figure}b).
This training was performed on a single NVIDIA V100 for 500 iterations with a batch size of 24 with images cropped to a resolution of $512 \times 512$
We used the same parameters as for pre-training, except that we changed the learning rate to $10^{-5}$.
Using this setup each fine-tune takes about 5 to 7 minutes.

\subsection{Classification Model}

For training the AMM, we first extracted the contrasts of all annotated flakes, yielding a contrast distribution.
We then denoised the distribution by applying a K-nearest neighbor classifier to all classes.
Afterwards, we applied DBSCAN~\cite{Ester1996} to remove any outlier points and normalized the distribution by setting the mean to zero and the standard deviation to one.
Finally, we sampled the points so that each class is represented by the same number of samples to counteract class imbalance.

For training, we used the Adam optimizer~\cite{Kingma2015} with a learning rate of 0.01, a batch size of 10000, and 5000 iterations with a dropout probability of $10\%$.
For the loss function, we used the standard cross entropy.
The network used an embedding dimension of $16$, a depth of $4$ and a spectral coefficient of $0.5$.
Training on a CPU takes about 5 minutes.

\section{Evalution}

\begin{figure*}[!bth]
    \begin{center}
    \includegraphics[width=\linewidth]{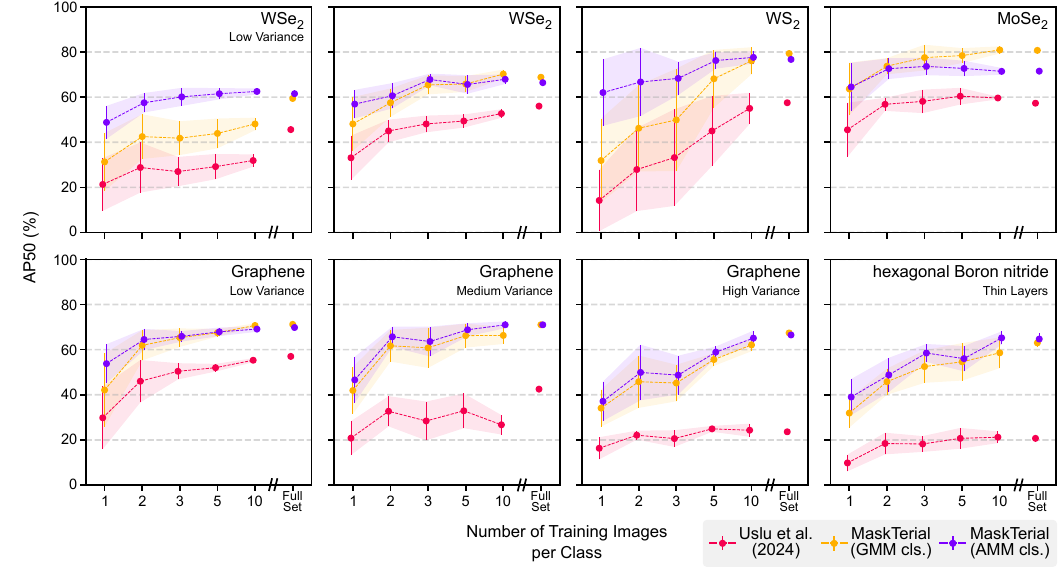}
        \caption{
The model has been evaluated on eight different datasets when trained with different amounts of images per class.
MaskTerial outperforms the baseline model from ref.~\cite{Uslu2024} for all training thresholds and for all materials.
An interesting find is that MaskTerial outperforms the fully trained baseline model with as little as two images per class for all materials.
Furthermore MaskTerial seems to saturate after as little as two images per class for most materials making further training unnecessary.
}
        \label{fig:metrics}
    \end{center}
\end{figure*}

We use the Average Precision at 50\% IoU (AP50) as the evaluation metric.
The AP50 is the area under the precision-recall curve at a threshold where the Intersection over Union (IoU) between predicted and ground truth boxes is at least 50\%.
This effectively measures the models performance in detecting instances of interest while minimizing false positives and is used as one of the default metrics when it comes to instance segmentation models.
The model was trained with varying numbers of training images to evaluate its performance and ability to handle few-shot learning tasks across different materials.

\subsection{Quantitative results}

We compare the Mask2Former instance segmentation model with an AMM classification head (MT-AMM) against the Gaussian mixture model (GMM) of ref.~\cite{Uslu2024} and the Mask2Former instance segmentation model with a GMM classification head (MT-GMM).
The models were evaluated for both the few-shot and full-data tasks with our eight datasets.
All models were trained and evaluated ten times in different data subsets to obtain metrics on their performance (see Figure \ref{fig:metrics}).

The results show that the MT-AMM and MT-GMM outperform the GMM baseline by a large margin of at least 10\% on all datasets and for any number of training examples.
The performance increase is particularly strong for materials with low optical contrast, such as thin layers of hBN, and highly varying substrate thicknesses, with metrics improving by up to 40\%.

For almost all materials, we see diminishing returns for the metrics even with more training examples, indicating that the model has already learned the distributions from only 2 to 5 example images.
The MT-AMM is particularly strong when in the low data regime, it outperforms the MT-GMM in this regime while also providing more stable performance (see Table \ref{tab:ablations})
When using more data, the MT-GMM starts to match the performance of the MT-AMM.

\subsection{Ablations}

\begin{table}[]
    \renewcommand{\arraystretch}{1.4}
    \centering
    \begin{tabularx}{\linewidth}{
    >{\centering\arraybackslash}X
   >{\centering\arraybackslash}X
   >{\centering\arraybackslash}X
  >{\centering\arraybackslash}X }
    \toprule
    \multirow{2}{*}{Model} & Synthetic pre-training & Classification Model & Average AP50 \\
    \midrule
    GMM Only & \xmark & GMM & 40.8 ± 15.5\\
    AMM Only & \xmark & AMM & 43.8 ± 12.2\\
    \midrule
    \multirow{6}{*}{MaskTerial}
    & \xmark   & Mask2Former   & 3.6 ± 3.5\\
    & \xmark   & GMM & 2.7 ± 2.8\\
    & \xmark  & AMM & 2.5 ± 2.4\\
    & \cmark & Mask2Former   & 35.2 ± 8.1\\
    & \cmark & GMM & 66.8 ± 10.6\\
    & \cmark & AMM & \textbf{68.9 ± 4.9}\\
    \bottomrule
    \end{tabularx}
\caption{
The table shows the impact of synthetic pre-training and classification model choice on the average AP50 scores of various model configurations. The deep learning instance prediction models without synthetic pre-training struggle significantly with detection, as shown by their low AP50 scores. Although using AMM as the classification model has a smaller impact on the AP50 than pre-training, it contributes to more stable and consistent results across datasets. The results are averaged across all datasets, with a threshold of 10 images per class.
}
\label{tab:ablations}
\end{table}

To determine which contributions most improved the model's performance, we conducted ablation studies.
Specifically, we evaluated the impact of synthetic pre-training and the choice of classification model (GMM vs. AMM vs. Mask2Former) on the detection metrics.

Table \ref{tab:ablations} highlights the effect of these components: models without synthetic pre-training achieve very low AP50 scores.
Having the Mask2Former model predict the classes itself also reduces performance, resulting in unstable and inconsistent detection across datasets.
These results show the critical role of synthetic pre-training and the stability advantage provided by AMM.

\section{Conclusion}

In this paper, we have presented a deep learning architecture, paired with a synthetic data generator, that improves upon existing algorithms for the detection of 2D material flakes in microscopy images, particularly for materials with low optical contrast.
Our model combines an instance prediction model with an uncertainty estimation model to make decisions based on physical features.
We have shown that our model significantly outperforms current state-of-the-art methods and can be trained on as few as 5 to 10 images per class in a few minutes.

The strength of our approach is the use of physical inductive biases in the model architecture.
By incorporating physical knowledge into the decision-making process, our model provides interpretable predictions that can be validated for further downstream processing, such as stacking of different 2D flakes into van der Waals heterostructures.
In addition, our few-shot learning capability allows for the detection of difficult to exfoliate and detect materials, which is a significant advancement over existing methods that require large annotated datasets.

We have also also introduced a synthetic data generator that mimics the true distributions of microscopy images.
This generator allows us to create large datasets for pre-training deep learning models, reducing the need for extensive data collection and annotation.
Our study shows that pre-training with synthetic data significantly improves the performance of our instance detection model, highlighting the potential of this approach in the 2D materials community.

Despite the strengths of our approach, there are some limitations to consider.
First, the model is less efficient at detecting small instances with an area less than 200 pixels.
This is a common challenge for deep learning models without specialized layers and techniques.
Second, when predicting instances that are close together, the model tends to combine them into a single prediction, leading to instance misclassification.

In summary, our novel deep learning architecture and synthetic data generator represent a significant step forward in the automated detection of 2D materials in microscopy images.
By exploiting physical inductive biases and few-shot learning capabilities, our models enable the detection of rare materials and provide interpretable predictions.
Although there are some limitations to our approach, we believe that our contributions lay the foundations for future research in this area and have the potential to have a mayor impact on the field of 2D materials science.

\section{Acknowledgments}
This project has received funding from the European Research Council (ERC) under grant agreement No. 820254 and the Deutsche Forschungsgemeinschaft (DFG, German Research Foundation) under Germany’s Excellence Strategy - Cluster of Excellence Matter and Light for Quantum Computing (ML4Q) EXC 2004/1 - 390534769.
A.N.\ acknowledges funding by the BMBF project ``WestAI'' (grant no.\ 01IS22094D).

\section{Data availability}
The code and data supporting the findings is hosted on Zenodo~\cite{Zenodo}.
The code and data for the model is also accessible on GitHub~\cite{Code}.
A demo website showcasing the synthetic data generator can be accessed at~\cite{Website}.

%%TC:ignore
%apsrev4-2.bst 2019-01-14 (MD) hand-edited version of apsrev4-1.bst
%Control: key (0)
%Control: author (8) initials jnrlst
%Control: editor formatted (1) identically to author
%Control: production of article title (0) allowed
%Control: page (0) single
%Control: year (1) truncated
%Control: production of eprint (0) enabled
%

%%TC:endignore

\end{document}